# Vision-based Warning System for Maintenance Personnel on Short-Term Roadwork Site

**Xiao Ni[1*], Walpola Layantha Perera[1], Carsten Kühnel[1], Christian Vollrath[1]**

1. University of Applied Sciences, Altonaer Straße 25, 99085 Erfurt, Germany

*xiao.ni@fh-erfurt.de

**Abstract:** We propose a vision-based warning system for the maintenance personnel working on short-term construction sites. Traditional solutions use passive protection, like setting up traffic cones, safety beacons, or even nothing. However, such methods cannot function as physical safety barriers to separate working areas from used lanes. In contrast, our system provides active protection, leveraging acoustic and visual warning signals to help road workers be cautious of approaching vehicles before they pass the working area. To decrease too many warnings to relieve a disturbance of road workers, we implemented our traffic flow check algorithm, by which about 80% of the useless notices can be filtered. We conduct the evaluations in laboratory conditions and the real world, proving our system's applicability and reliability.

## 1. Introduction

The safety of road maintenance workers has been a well-known object of traffic research and development projects in Europe in recent years. Guidelines for human behavior, roadwork site setups, and technical solutions were developed and implemented [1, 2, 3]. Almost all of these guidelines deal with preventing rear-end collisions with safety trailers. Another issue that has not been dealt with yet is the safety of maintenance personnel, especially their safety within the area of the short-term roadwork sites (STRWS). When the maintenance workers are working, sometimes they cannot sense the surroundings explicitly. Moreover, the construction vehicles in the front can obscure the view of the road workers, and they cannot see the traffic behind them.

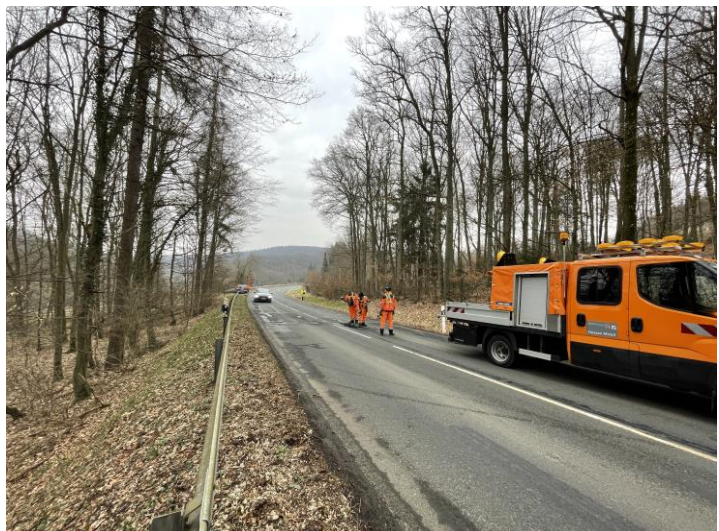

**Figure 1: Short-term roadwork site**

In many European countries, the short-term roadwork sites are separated from lanes with free-flowing traffic only by mobile warning signs such as traffic cones, safety beacons, or even nothing (Figure 1). Protection by passive protective devices is not possible due to the limited duration, and no physical barriers can hold back vehicles in case of an accident. As a result, personnel in STRWS are generally exposed to higher risks. Therefore, the German Federal Ministry of Labour and Social Affairs implemented a guideline to lower the risk of accidents in short-term roadwork sites. However, Hands-on experience shows that this is difficult to observe for maintenance workers, as there are many situations in which the "No-Entry Zone" is entered, consciously or unconsciously.

Our solution addresses this problem by trying to detect the approaching vehicle from both front and rear directions based on computer vision and trigger a warning signal to alert the maintenance personnel. To reach that aim, we utilize the recent groundbreaking deep learning-based Object detection for detecting approaching vehicles on the road in both directions. Furthermore, the system can track such vehicles and record their trajectories. After tracking vehicles, we utilize our traffic flow check algorithm to filter unnecessary warnings.

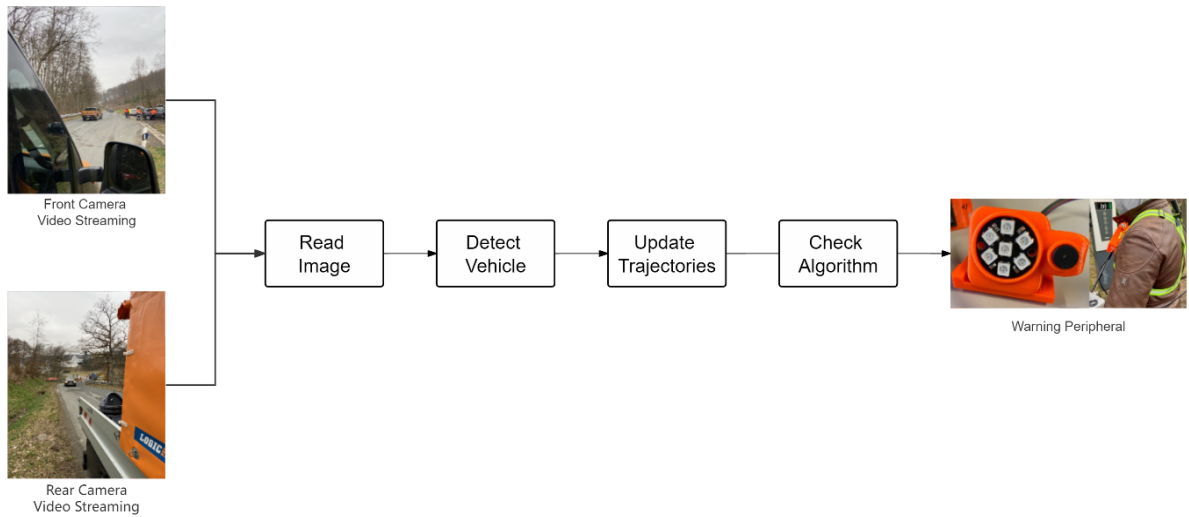

**Figure 2: Architecture of the vision-based warning system**

There are two inputs to the system. One is a video streaming of a front-mounted conventional RGB camera, and another is a video streaming of the rear-mounted conventional RGB camera. With these two cameras, we can observe both directions of the construction site. The whole system is composed of three main components: (1) a fast detector, (2) a detector-based tracker (3) a traffic flow check algorithm (see Figure 2).

## 2. Related works

**Vehicle detection** is a considerably mature and proven technology that is crucial for the whole system. Traditional detectors combine the sliding window method with traditional machine learning classifiers [4, 5]. Such methods rely on handcrafted features, require much feature engineering work, and cannot achieve satisfactory accuracy. In recent decades, due to the groundbreaking improvements in deep learning, modern detectors are mainly composed of the convolutional neural network (CNN) [6, 7, 8]. CNN has a similar property to the traditional fully connected network but considerably reduces the amount of the neural network's parameter, making CNN much more robust and less prone to overfitting. Such CNN-based detectors require large-scale datasets to train their neural network and can achieve a better generalization capability. As a result, their accuracy is significantly improved in comparison to traditional detectors. Since the working condition of maintenance is outdoor and constantly varies from one country road to another, some country roads' backgrounds are dense forest and mountain bodies, while the others ought to be open fields. The weather can also vary seasonally, from rain to sunshine. Therefore, we need a robust and highly reliable detector for the application and choose a CNN-based detector. CNNs have a high requirement of computing resources to get a fast inference speed, so in the project, we equipped the vehicle computer with GPU, which can accelerate the neural network's inference.

There are also two main branches of modern detectors. The first is R-CNNs based on the regional proposal, and the inference process consists of two stages [6, 9, 10]. The other is a one-shot method, which operates the feature extraction and object localization at the same time [7, 8, 11]. The one-shot methods have a relatively lower accuracy but much less inference time. Since it is a real-time application with a high running time requirement, we decided to use a type one-shot detector. In recent years, **multiple object tracking** (MOT) has gained increasing attention, and there are still many challenging tasks like severe object occlusion and abrupt appearance changes [12]. We use MOT to find out multiple objects in single frames and calculate the trajectory of each object across continuous frames. According to the criterion of initialization method, MOT can be categorized into Detection-based Tracking (DBT) [13, 14] and Detection-Free Tracking (DFT) [15, 16]. DFT has a manually defined and fixed number of objects, so it cannot handle the new appearance and disappearance of objects and is inappropriate for our application. In contrast, DBT uses detectors to discover new objects without number limitations, and disappearing objects are abandoned automatically.

MOT can also be grouped into offline tracking [14, 17] and online tracking [18]. Offline methods utilize batches of frames or the entire image sequence to process the data. Nevertheless, our real-time application requires low inference time, so online methods would be more appropriate. We can predict the object's current state solely based on the observations from the past up to the current frame.

### 3. Vision-based warning system
In the following, we will describe the hardware components that are used in our application (Section 3.1) and discuss the proposed detector (Section 3.2) and detection-based tracking (Section 3.3) in more detail. Section 3.3 then explains the traffic flow check algorithm.

### 3.1 Hardware
The components of the system will include two traditional RGB cameras mounted on the maintenance vehicle and a computer with the graphic processing unit GeForce GTX 1080 Ti for computer vision purposes (See Figure 3). The vehicle computer can be operated directly using a 12 Volt power supply from the vehicle's electrical system, which means that it can be operated directly in the vehicle without the need for major structural modifications. All that requires is an adequately protected and dimensioned connection to the vehicle's electrical system. Figure 3 displays the computer which was purchased for our application. The operating system is Linux-based and thus offers a good platform for vision-based applications.

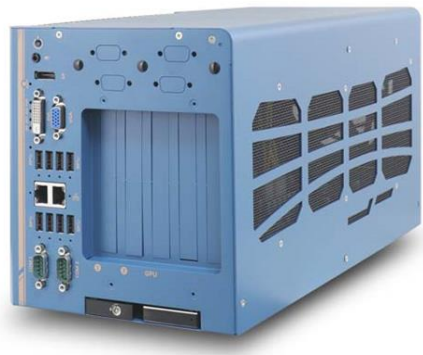

**Figure 3: Nuvo-8108GC-XL (Quelle: www.neousys-tech.com)**

The peripheral component of the system is a warning signal device worn by road maintenance workers. The component can be attached to a chest strap to improve wearing comfort (Figure 4). The acoustic signal is implemented by a piezo buzzer, as in smoke alarms. The reason for this choice is that the frequency of the hearing protector does not filter these frequencies strongly in the buzzer range, and the signals can be accordingly easily perceived by the road workers. The acoustic signal generator, in combination with the visual warning lights, is attached to the chest strap.

### 3.2 Approaching vehicle detection
In the first step, we need to detect the approaching vehicles in each frame of both camera streams. We select YOLOv4 to perform the detection. In the following, we describe the training phase of YOLO.

There are three classes in our application to be categorized. They are trucks, vehicles, and pedestrians. We used the pre-trained weights from the original author[1] and applied the transfer learning concept. The transfer learning concept means the knowledge that the neural network has learned from a previous task could be applied to the current task [19]. Transfer learning can dramatically reduce the time required to train neural networks from scratch, and the model can yet perform similarly. Here, we fine-tuned the neural network with our dataset to adapt the neural network to our application.

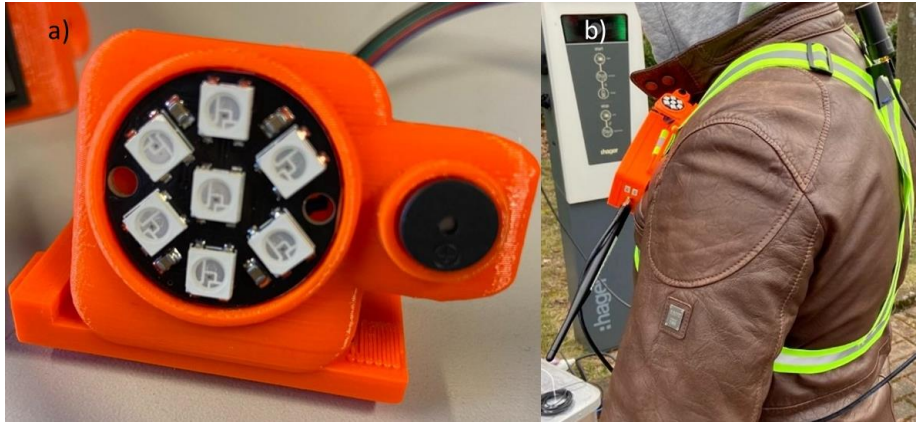

**Figure 4: Personal Road Worker Device**

We have created a dataset based on the videos taken from actual maintenance environments. Our dataset contains 1200 images with annotations. The project team members manually performed the labeling task for the dataset. Moreover, the dataset was randomly split into three sets (training/development/test), where the allocation percentage of these three sets follows the rule of thumb (60/20/20).

Since we had a relatively small training set, we solely deleted the output layer and the weights feeding into that layer. Then, we created a new output layer with the output tensor shape $N \times N \times [3 \times (4 + 1 + 3)]$. $N \times N$ represents the number of grid cells, and 3 is the number of anchor boxes. The bounding box of each anchor box is defined by $(c_n^i, w_n^i, h_n^i), n \in [1, N \times N \times 3]$, where $c_n^i = (x_n^i, y_n^i)$ represents the center, $w_n^i$ the width and $h_n^i$ the height of the corresponding box. $i$ implies the frame index. 1 Objectness $C_n^i$ denotes the probability that an object belonging to any class in the training set exists in the bounding box. 3 class confidence scores $P_n^{i(j)}(C_n^i), j \in [1, 3]$ are the conditional probability, i.e., probability of class x given an object exists in this box. The Products of the Objectness and the class confidence scores specify the probability that the bounding box contains a specific object type.

The weights of the output layer were randomly initialized. Only the weights of the last two layers were optimized, and other layers of the neural network were fixed because of the lack of a large-scale training dataset and the corresponding overfitting concern. The learning rate was set to 0.001, ten times lower than the regular learning rate. The neural network was trained for 20 epochs.

### 3.3 Detection-based tracking
Since the maintenance workers work in the outdoor environment, imaging conditions make a significant difference to the visual appearance of the approaching vehicles, which depends considerably on weather conditions, sunlight intensity, and light reflection and varies significantly over time and location. For this reason, we do not apply the visual appearance model but solely rely on the motion model to relate the detections to the trajectories, similar to [20].

We applied a Kalman filter as the probabilistic inference model to represent the states of each approaching vehicle. Here we assume the approaching vehicle travels at a constant velocity since the vehicle's acceleration is not that high, and the interval between image frames is sufficiently small. Our test also proves that the higher-order motion assumption-based model did not improve the precision of the vehicles' location. Therefore, we employed the Kalman filter model with the assumption of constant velocity to increase computational efficiency.

We define the Euclidean distance $\left\|\hat{c}_k^i - c_n^i\right\|_2$ as the cost of detection being assigned to a trajectory $t_k$, where $\hat{c}_k^i = F\left(c_k^{i-1}, \; v_k^{i-1}\right)$ is the filter's state estimation of the trajectory at frame $i$. Here, $c_k^{i-1}$ is the location of the $k$-th vehicles at the previous frame $i-1$, $v_k^{i-1}$ the velocity, and $F$ the state transition function. After calculating the matching cost, the Hungarian algorithm is employed to solve the optimal assignment of detection-to-trajectory [21, 22]. Therefore, we update the trajectory of the $k$-th vehicles with $t_k^i = t_k^{i-1} \cup c_n^i$ in case new detection $c_n^i$ is assigned to the trajectory $t_k^{i-1}$. Then, the filter's state is updated with the new observation.

If some detections cannot be assigned to existing trajectories, new trajectories are tentatively initialized for them, and such trajectories stay suspended. The trajectories could be set as active only if detections could be assigned to them in the subsequent frames. They will be terminated if the trajectories cannot match any detections for several consecutive frames.

### 3.4 Traffic flow check algorithm

Every time a vehicle is tracked with the help of the detection-based tracking system, the maintenance workers will be not only acoustic but also visual warned. However, in one peak hour, hundreds of vehicles approach the construction site, and for each vehicle, the warning device is triggered once, and the road worker is alerted. That means the road workers should be warned hundreds of times in one hour. Apparently, it is not an appropriate solution for road workers. Road workers cannot endure being constantly distracted from their work by such warnings, and such notices can even cause light and sound contamination to them. To overcome these issues, we also developed a traffic flow check algorithm to filter such identification signals from the tracking system.

The basic concept of the whole warning system is to alert the maintenance workers of the coming vehicles when they might be unaware of the possible danger. When a car approaches and the system alerts the road workers, they will already prepare and start observing the road condition. At that time, another vehicle is driving to the construction site. There is no need to alert the road workers again because they have already been on guard for the passing cars. The traffic flow will also keep the road workers paying attention to the road condition. It is necessary to alert the road workers of the newly emerging vehicle only if there are no vehicles on the road and the road workers relax their attention.

---

**Algorithm 1** Traffic Flow Check Algorithm

**Begin**
    $T_{duration} := 10$
    Set $t_{start}$ as the current timestamp
    **While True**
        **If** a new vehicle is identified **Then**
            Set $t_{end}$ as the current timestamp
            $t_{diff} = t_{end} - t_{start}$
            **If** $t_{diff} > T_{duration}$ **Then**
                Trigger Warning
            **End If**
            Set $t_{start}$ as the current timestamp
        **End If**
    **End While**
**End**

---

The traffic flow check algorithm uses a while loop to monitor the signal from the tracking system. The tracking system will signal the check algorithm when it identifies a new vehicle. After receiving the signal, the check algorithm will check whether the elapsed time from the last signal to this one $t_{diff}$ exceeds the pre-defined time duration $T_{duration}$. If true, the check algorithm will eventually trigger a warning alerting road workers. If false, the start time $t_{start}$ will be reset as the current timestamp, and the timing process will restart. In the initialization phase, $t_{start}$ will be set as the timestamp at that moment. With the algorithm, we can ensure that the check algorithm can trigger a warning only if there is no traffic on the country road for 10 seconds since the last vehicle. The setting of the $T_{duration}$ as 10 seconds is the tradeoff between convenience and safety. Despite necessary warnings, the road workers should be minimal disturbed. Empirical evidence shows that maintenance workers can maintain the warning caused attention for about 10 seconds.

## 4. Evaluation

### 4.1 Detector performance

Both accuracy and the inference time of the underlying detectors are key performance Indicators for the whole system. The detector should accurately identify the objects at a high frame rate. A high frame rate means that the elapsed time between frames is short; therefore, the locations change between frames is small, which ensures reliable matching between detections and trajectories. The detector will be evaluated by the test set of the application-specific dataset, which covers a diverse collection of images taken in different weather and sunlight conditions by the two cameras, which are mounted on the vehicle. Therefore, the evaluation with this test set can demonstrate how well the detector can perform in actual working environments. The average precision of the detector can reach 96.2 %, while the runtime can still be restricted to 26.9 $ms$, which equals 37.2 frames per second (FPS). The frequency of the image input of cameras is 30 FPS, which is lower than the processing speed of the detector. That means the detector can be applied to the real-time application in combination with the two cameras. The bottleneck lays now not on the detector but the sensor.

### 4.2 Real-world test

The system's effectiveness is not only in laboratory conditions but also in the real world evaluated. For example, we installed the whole system in a construction vehicle, and two maintenance road workers wore the corresponding warning peripherals. During the testing days, the road workers solely did the regular work, like cleaning potholes, inserting asphalt, compressing, etc. Throughout the long-term study, we recorded (1) the timestamps of warning the road workers and (2) the timestamps of the vehicles which caused the corresponding warning passing the construction site. Then, we can calculate the time difference between both timestamps and count the number of time differences that belong to the fixed range.

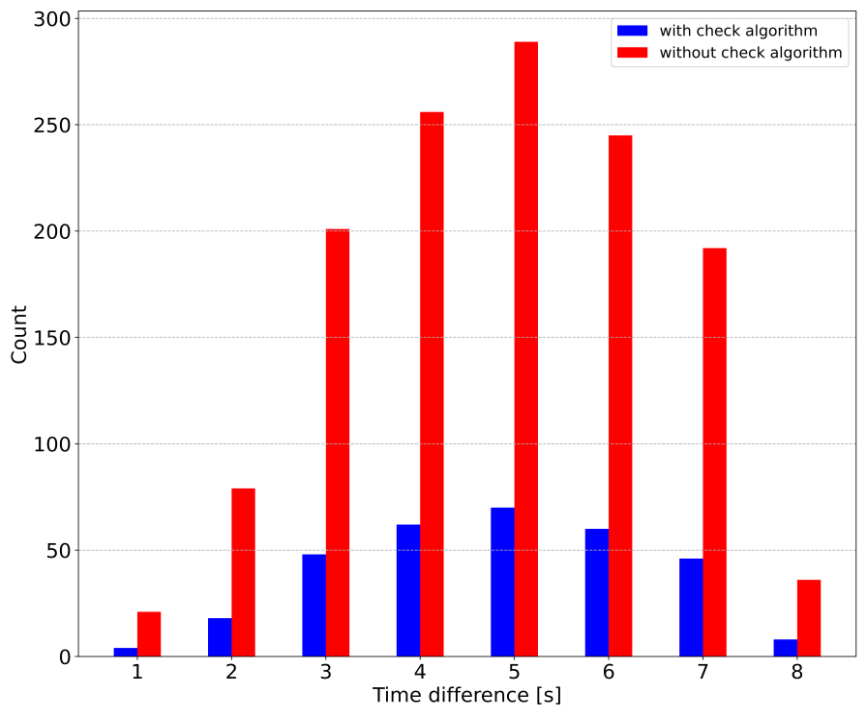

**Figure 5: Time difference between warning of the road workers and vehicles passing the construction site and the count of the warnings in one day (with and without traffic flow check algorithm)**

Figure 5 demonstrates the whole system's performance: How many seconds before the approaching vehicles pass the construction site can the detector recognize the vehicle and alert the road workers by the visual and acoustic signals of the warning device? The pre-warning duration represents the remaining response time for the road works and is crucial to the performance of our application. From the figure, we can see that our system can reliably report the warning 3 to 7 seconds before the approaching vehicle passes the construction site, which allows the road worker to have adequate time to prepare for the potential dangers. We performed a manual analysis for the cases where the time differences were only equal to about 1 to 2 seconds. Such cases happened when the construction site was in the middle of the curve of some country road. The camera's view will be blocked by trees or mountain bodies so that it can't detect the approaching vehicle earlier. But this scenario itself is even more dangerous without our warning system. With the help of the front camera on the car and the vehicle recognition system, we can win up to 2 seconds for the road workers to prepare, which significantly improves the safety of construction in this dangerous scenario.

In addition, Figure 5 can present the influence of the implementation of the traffic flow check algorithm. From the graph, we can count the number of alerts received by the road worker in one day in the condition with or without the check algorithm. The road workers receive 1308 warnings in one day, on average 164 warnings per hour without the check algorithm. In peak hours, the road worker can receive even about 32 warnings per minute, which is unacceptable for the road workers. When implementing the check algorithm, the warning times are filtered to one-fifth of the original times, only 316 times. And especially in peak hours, the check algorithm makes a big difference. The received warning times are restricted to only about one warning per minute when the traffic flow is very dense because the vehicles that follow the first approaching vehicles are very close. The duration between detecting those vehicles is usually lower than $T_{duration}$ (10 seconds) and will trigger no warnings. In such a situation, the road workers are aware of the existence of such approaching vehicles, so there is no need to remind them repeatedly.

On the one hand, the system can help road workers detect approaching vehicles when they focus on work and are not so sensitive to the environment; on the other hand, it can give warning to the road workers in advance when approaching vehicles are still obscured by the construction vehicle, which results in that the road workers cannot detect such vehicles in time. The latter is extremely dangerous and can be thoroughly solved by the vision-based warning system.

## 5. Conclusion

The safety of road maintenance workers on STRWS is critical, mainly because there are no physical safety barriers separating working areas and used lanes. Instead, the roadwork site and the traffic lane are separated only by traffic cones, safety beacons, or even nothing. The approach proposed in this paper, which consists of a detector, tracking, check algorithm, and warning device, aim to improve the safety of maintenance workers in STRWS with active protection. The evaluations demonstrate that the implemented system can reliably detect the approaching vehicle and decide intelligently whether the warning device should alert the road workers or not. With two cameras in both the front and rear of the construction vehicle, both sides of the country road will be under observation. The developed traffic flow algorithm can efficiently reduce the warning times so that the road workers are not too often disturbed and receive just necessary warnings. Furthermore, the warning device can send both visual and acoustic signals. The acoustic signal is implemented by a piezo buzzer, whose frequency will not be significantly filtered by the hearing protector that the road worker wears during the work to filter the noise from the roads.

## 6. Acknowledgement

The work described in this paper has been funded by the German Federal Ministry of Transport and Digital Infrastructure within the project MOSAik:D (Grant number 01MM19006C).

## 7. References

1. Hessen Mobil - Road and Traffic Management. (2018). *Handbuch zum Baustellenmanagement*. Hessen Mobil, [Online]. Available: https://mobil.hessen.de/sites/mobil.hessen.de/files/Hessen_Mobil_Baustellenmanagement_2018.pdf [Accessed 07 January 2021] (in German).
2. Hessen Mobil - Road and Traffic Management. (2018). *Hessischer Verkehrszeichenplankatalog für Arbeitsstellen kürzer Dauer (HE-VZP-Katalog*. Hessen Mobil [Online]. Available:

https://mobil.hessen.de/sites/mobil.hessen.de/files/content-downloads/HE%20VZP-Katalog_AkD_StandAugust2016.pdf [Accessed 07 January 2021] (in German).

3. German Ministry of Traffic and Digital Infrastructure. (2020). *Cooperative Traffic Systems –safe and intelligent*. Cooperative ITS corridor, [Online]. Available: https://c-its-korridor.de/?menuId=1&sp=en. [Accessed 07 January 2021] (in German).

4. L. Mao, M. Xie, Y. Huang and Y. Zhang, "Preceding vehicle detection using Histograms of Oriented Gradients," 2010 International Conference on Communications, Circuits and Systems (ICCCAS), 2010, pp. 354-358, doi: 10.1109/ICCCAS.2010.5581983.

5. D. R. Sulistyaningrum, T. Ummah, B. Setiyono, D. B. Utomo, Soetrisno, B. A. Sanjoyo. Vehicle detection using histogram of oriented gradients and real adaboost, Journal of Physics: Conf. Ser. 1490 012001

6. He, K., Gkioxari, G., Dollar, P., Girshick, R. (2017). Mask R-CNN. in International Conference on Computer Vision (ICCV), Venice.

7. W. Liu, D. Anguelov, D. Erhan, C. Szegedy, S. Reed, C. Y. Fu, A. C. Berg. SSD: Single Shot MultiBox Detector. In Proc. ECCV, 2016

8. J. Redmon and A. Farhadi. YOLO9000: Better, Faster, Stronger. In Proc. CVPR, 2017.

9. S. Ren, K. He, R. Girshick, and J. Sun. Faster R-CNN: Towards Real-Time Object Detection with Region ProposalNetworks. In Proc. NIPS, 2015.

10. Z. Cai and N. Vasconcelos, "Cascade R-CNN: delving into high quality object detection," Proceedings of the IEEE Conference on Computer Vision and Pattern Recognition, pp. 6154–6162, 2018.

11. J. Redmon and A. Farhadi, "Yolov3: an incremental improvement," arXiv:1804.02767, 2018.

12. W. Luo, J. Xing, A. Milan, X. Zhang, W. Liu, X. Zhao, andT.-K. Kim. Multiple Object Tracking: A Literature Review. arXiv CoRR, abs/1409.7618, 2017.

13. B. Bose, X. Wang, E. Grimson, Multi-class object tracking algorithm that handles fragmentation and grouping, CVPR, 2007.

14. B. Song, T.-Y. Jeng, E. Staudt, A. K. Roy-Chowdhury, A stochastic graph evolution framework for robust multi-target tracking, ECCV, 2010.

15. L. Zhang, L. van der Maaten, Structure preserving object tracking, CVPR, 2013.

16. M. Yang, T. Yu, Y. Wu, Game-theoretic multiple target tracking, ICCV, 2007.

17. Z. Qin, C. R. Shelton, Improving multi-target tracking via social grouping, CVPR, 2012.

18. C.-H. Kuo, C. Huang, R. Nevatia, Multi-target tracking by online learned discriminative appearance models, CVPR, 2010.

19. N. Gonthier, Y. Gousseau, S. Ladjal, An analysis of the transfer learning of convolutional neural networks for artistic images, ICPR Workshop, 2020

20. H. Possegger, T. Mauthner, P. M. Roth, and H. Bischof. Occlusion Geodesics for Online Multi-Object Tracking. In Proc. CVPR, 2014.

21. Z. Qin, C. R. Shelton, Improving multi-target tracking via social grouping, CVPR, 2012.

22. V. Reilly, H. Idrees, M. Shah, Detection and tracking of large number of targets in wide area surveillance, ECCV, 2010.